# Fiber Transmission Model with Parameterized Inputs based on GPT-PINN Neural Network

YUBIN ZANG,[1] BOYU HUA,[2] ZHIPENG LIN,[2] FANGZHENG ZHANG,[2] SIMIN LI,[2] ZUXING ZHANG[1,*] AND HONGWEI CHEN[3,*]

[1]*College of Electronic and Optical Engineering, Nanjing University of Posts and Telecommunications, Nanjing, China, 210023.*
[2]*College of Electronic and Information Engineering, Nanjing University of Aeronautics and Astronautics, Nanjing, China, 211106.*
[3]*Department of Electronic Engineering, Tsinghua University and Beijing National Research Center for Information Science and Technology, Beijing, China, 100084.*
*\*zxzhang@njupt.edu.cn   \*chenhw@tsinghua.edu.cn*

**Abstract:** In this manuscript, a novelty principle driven fiber transmission model for short-distance transmission with parameterized inputs is put forward. By taking into the account of the previously proposed principle driven fiber model, the reduced basis expansion method and transforming the parameterized inputs into parameterized coefficients of the Nonlinear Schrödinger Equations, universal solutions with respect to inputs corresponding to different bit rates can all be obtained without the need of re-training the whole model. This model, once adopted, can have prominent advantages in both computation efficiency and physical background. Besides, this model can still be effectively trained without the needs of transmitted signals collected in advance. Tasks of on-off keying signals with bit rates ranging from 2Gbps to 50Gbps are adopted to demonstrate the fidelity of the model.

## 1. Introduction

Nowadays, the disciplinary field of artificial intelligence (AI) and fiber optics has been developing at a remarkable pace[1-3]. When it comes to fiber transmission, the characteristics of how the attenuation, dispersion and non-linearty have influence on the signal propagation can be accurately described by the Non-linear Schrödinger Equation (NLSE)[4]. Like other partial derivative equations, NLSE rarely posses analytical solutions. Therefore, traditional communication simulation software usually adopts the Split Step Fourier Method (SSFM) to obtain the numerical solutions. By dividing the whole transmission distance into various basic computing units and assuming that the effect of dispersion and non-linearity decouples in each computing unit, SSFM can sequentially compute the signal waveforms unit by unit till the end [5-6]. Though SSFM is accurate as long as the length of computing unit is appropriately determined, it may consume large scale of running time especially for cases with long distance or strong non-linearity.

Researchers have been trying to utilize the AI's strong regression ability to overcome the drawback of SSFM. Indeed, multiple AI based fiber transmission models have been put forward in recent 10 years. Among them are bidirectional long short time memory (BiLSTM) models [7], Generative Adversarial Network (GAN) [8-9] models, attention mechanism models [10-11] and etc. These models have greatly increased the computational efficiency but discard the physical background behind the signal transmission since they all view as the pure regression tasks. Besides, these models can be ineffective unless the transmitted signals are collected in advance. In order to better keep the balance between computing efficiency and the physical background, principle driven fiber models were proposed. By adopting NLSE as its loss functions to guide the model's convergence, the model can progressively establish the relations between the coordinates of both time and distance with the shape of signals. The model can not only be effectively trained without the pre-collected signals after transmission, but also remain the physical background. However, for the optimizing of fiber communication systems which in important in the B5G communication system establishing, the inputs are always changeable with various bit rates, but it is inconvenient for the principle driven fiber transmission model to efficiently compute because it needs to be retrained for different input signals [12-13].

Therefore, In this manuscript, the generalized principle driven fiber transmission model which can deal with input signals with different bit rates without re-training the whole model is proposed. This work can be successfully completed by transforming the parameters of the inputs into the coefficients of NLSE and taking both previously proposed principle fiber driven model and the reduced basis expansion method to establish the generalized model. The article will be developed into four parts. Backgrounds and motivations will be illustrated in the introduction part. Then both principles of parameter transform, principle driven models and reduced basis method will be developed in the second part. Task of OOK with bit rates ranging from 2Gbps to 100Gbps which is utilized to test the whole model will be illustrated and analyzed in the part of simulation setups and analysis. Advantages and further research interests will be discussed in the conclusion part.

## 2. Model structures and configurations

### 2.1 Fiber Transmission and Parameter Transform

According to the theory of fiber optics and transmissions, Nonlinear Schrödinger Equation (NLSE) is used to mathematically describe how attenuation, dispersion and non-linearity affect the signal propagation which can be expressed as

$$\begin{cases} i\dfrac{\partial S}{\partial z} + i\dfrac{\alpha}{2}S + \beta_2 \dfrac{\partial^2 S}{\partial T^2} + i\beta_3 \dfrac{\partial^3 S}{\partial T^3} + \gamma |S|^2 S = 0 \\ S(z=0, T; R_b) = F(T; R_b) \end{cases} \quad (1)$$

in which $S = S(T, z; R_b)$ denotes the solution of NLSE under the initial condition $F(T; R_b)$. Here, both solution and initial condition closely related with bit rate, therefore, $R_b$ becomes the important parameter. $z$ and $T$ represents the original distance and time coordinate under the time-retarded frame. $\alpha$, $\beta_2$, $\beta_3$ and $\gamma$ is the loss per distance, second-order phase propagation constant, third-order phase propagation constant and nonlinear coefficient which reflect the attenuation, dispersion, high-order dispersion and nonlinear effect respectively.

Observed from Eq.(1), parameters are introduced by the initial condition which seems to be hard to handle. But fortunately, since only bit rate varies, this parameter can be moved from the initial condition into coefficients of NLSE by choosing the appropriate time coordinate transform. Other coordinate transform including time, distance coordinate transform, intensity transform are necessary so as to obtain the standard NLSE in which normalized time ranges from -1 to 1, normalized distance ranges from 0 to 1 and normalized amplitude or intensity of the signal ranges from 0 to 1. This normalization can also let the neural network based model which will be illustrated later work in a proper way. In all, the coordinate transforms are defined as

$$\begin{cases} \gamma = \dfrac{n_2 \omega_c}{c A_{eff}}, L_D = -\dfrac{1}{R_b^2 \beta_2}, L_{NL} = -\dfrac{1}{\gamma P_0} \\ z = L_{max}\zeta = \dfrac{L_{max}}{L_D} L_D \zeta = \kappa_1 L_D \zeta \\ T = T_{max} t = T_{max} R_b \dfrac{1}{R_b} t = \dfrac{\kappa_2 \zeta}{R_b} \\ S = \sqrt{P_0} s \end{cases} \quad (2)$$

in which $L_D$ and $L_{NL}$ represents dispersion length and nonlinear length respectively. $t$ and $\zeta$ represents the normalized time and distance coordinate under the time-retarded frame while $s=s(t,\zeta)$ represents the normalized solution whose intensity or amplitude ranges from 0 to 1. $P_0$ is the maximum power of the original signal. $n_2$, $\omega_c$, $c$ and $A_{eff}$ denotes nonlinear refractive index, central angular frequency of light, speed of light and effective cross-sectional area respectively. As can be seen from the time coordinate transform, the parameter $R_b$ is successfully transformed from the initial condition into the coefficient of NLSE. By substituting Eq.(2) into Eq.(1), the normalized NLSE with varying bit rate signal as its input can be expressed as

$$\begin{cases} iA_1 \dfrac{\partial s}{\partial \zeta} + i\kappa_1(R_b) A_2(R_b) s + \kappa_1(R_b) A_3 \dfrac{\partial^2 s}{\kappa_2^2(R_b)\partial t^2} \\ +i\kappa_1(R_b) A_4(R_b) \dfrac{\partial^3 s}{\kappa_2^3(R_b)\partial t^3} + \kappa_1(R_b) A_5(R_b)|s|^2 s = 0 \\ s(\zeta=0,t) = f(t) \end{cases} \quad (3)$$

in which $s=s(t,\zeta)$ represents the normalized solution. $R_b$ now becomes the parameter of those coefficients $\kappa_1$, $\kappa_2$, $A_2, A_4$, $A_5$ of the normalized NLSE. The initial condition $f(t)$ is no longer related with the symbol rate. Under this circumstance, those five equals

$$A_1 = 1 \quad A_2 = \dfrac{\alpha L_D}{2} = -\dfrac{\alpha}{2\beta_2 R_b^2} \quad A_3 = -\dfrac{sign(\beta_2)}{2}$$

$$A_4 = -\dfrac{\beta_3 L_D R_b^3}{6} = \dfrac{\beta_2 R_b}{6\beta_3} \quad A_5 = \dfrac{L_D}{L_{NL} R_b^2} = -\dfrac{\gamma P_0}{2\beta_2 R_b^2} \quad (4)$$

### 2.2 The Principle Driven Fiber Model

The principle driven fiber model as previously proposed is a multi-layer fully connected network with respect to normalized time and distance coordinates $(t,\zeta)$ under the time-retarded frame as its inputs and the real and imaginary part of the waveform value $(s_R,s_I)$ as its outputs. Linear multiplications and nonlinear $Tanh()$ activation consist of the basic computations between each layer.

Unlike other data driven models which utilize normalized mean square error (NMSE) as their loss functions, the principle driven fiber model substitutes its prediction into NLSE to calculate the residue error as to measure the prediction precision. Besides, the principle driven fiber model also compares the distance between the prediction at 0 distance with the initial condition, i.e, the input signal. This loss function can be defined as

$$L^F := \left\{ \dfrac{1}{N_E} \sum_{k=1}^{N_E} \left\| \begin{aligned} & iA_1 \dfrac{\partial s_k^{pred}}{\partial \zeta} + i\kappa_1 A_2 s_k^{pred} + \kappa_1 A_3 \dfrac{\partial^2 s_k^{pred}}{\kappa_2^2 \partial t^2} \\ & +i\kappa_1 A_4 \dfrac{\partial^3 s_k^{pred}}{\kappa_2^3 \partial t^3} + \kappa_1 A_5 \left|s_k^{pred}\right|^2 s_k^{pred} \end{aligned} \right\|^2 + \dfrac{1}{N_{ini}} \sum_{k=1}^{N_{ini}} \left\| \begin{aligned} & s_k^{pred}(\zeta=0,t) \\ & -s_k(\zeta=0,t) \end{aligned} \right\|^2 \right\} \quad (5)$$

In Eq.(5), $N_g$ and $N_{ini}$ represents the points of the solution and initial condition respectively.

### 2.3 Reduced Basis Expansion Method

Until now, the solutions for the previously proposed fiber model is a one-by-one solution which indicates that each model should be specifically trained for each $R_b$ which can consume large amounts of time when the scale of $R_b$ is large. In order to compress both the scale and number of fiber models to be trained to obtain the solutions for each $R_b$, reduced basis expansion method is utilized.

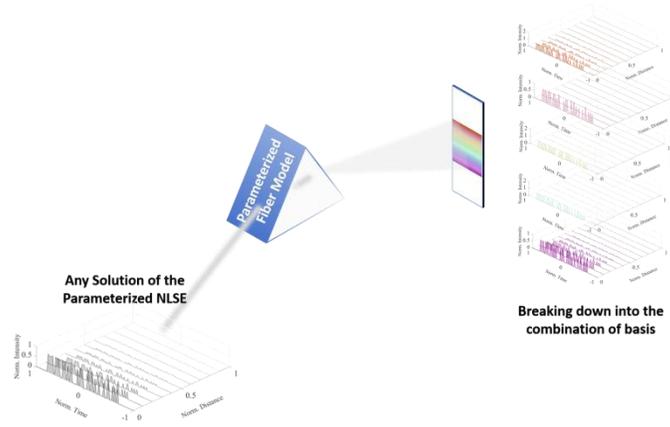

Fig. 1 Principle of parameterized fiber communication model

The basic idea behind the method comes from breaking the sophisticated generalized object into several main components like using a prism to breaking the sunlight into different components with different colors as is shown in Fig.1. Specifically, the reduced basis expansion method is to conduct the linear combinations of basis to approximate the universal solution [14]. Here, some 'specially chosen' solutions of NLSE for specific $R_b$ is performed as the basis to construct the final prediction results which can be expressed as

$$s_p^{pred} = \sum_{m=1}^{p} c_m s_m^{pred} = \sum_{m=1}^{p} c_m s_{Rm}^{pred} + i \sum_{m=1}^{p} c_m s_{Im}^{pred} \tag{6}$$

in which $c$ represents the linear combination coefficients and take real values.

The basic solutions in Eq.(6) and the linear combination coefficients can be both found by greedy algorithm. At the beginning, one value of $R_b^1$ is chosen. Then, the principle driven fiber model is trained to found the solution with respect to $R_b^1$. Afterwards, reduced basis expansion method will take into effect to use this solution to approximate other solutions by optimizing the linear combination coefficient. After finding the coefficient, different approximation errors for different $R_b$ can be calculated. Then the one with the largest approximation error becomes the second bit rate $R_b^2$ which will be used to train the principle driven fiber model to obtain the second basic solution. The same procedure repeats unless either the minimum approximation error or the maximum training epoch which is determined before the training is reached.

It should be noticed that, the findings of $c$ in greedy algorithm can adopt the gradient descend method in order to increase the search efficiency. In order to do so, it is necessary to calculate both values and gradients of approximation errors. The errors can be expressed as

$$L^{PF} := \left\{ \frac{1}{N_E} \sum_{k=1}^{N_E} \left\| \begin{array}{l} iA_1 \frac{\partial}{\partial \zeta} \sum_{p=1}^{N_n} c_p s_{p,k}^{pred} + i\kappa_1 A_2 \sum_{p=1}^{N_n} c_p s_{p,k}^{pred} \\ + \kappa_1 A_3 \frac{\partial^2}{\kappa_2^2 \partial t^2} \sum_{p=1}^{N_n} c_p s_{p,k}^{pred} + i\kappa_1 A_4 \frac{\partial^3}{\kappa_2^3 \partial t^3} \sum_{p=1}^{N_n} c_p s_{p,k}^{pred} \\ + \kappa_1 A_5 \left| \sum_{p=1}^{N_n} c_p s_{p,k}^{pred} \right|^2 \sum_{p=1}^{N_n} c_p s_{p,k}^{pred} \end{array} \right\|^2 + \frac{1}{N_{ini}} \sum_{k=1}^{N_{ini}} \left\| \sum_{p=1}^{N_n} c_p s_{p,k}^{pred} (\zeta=0,t) - s_k (\zeta=0,t) \right\|^2 \right\} \tag{7}$$

in which $N_n$ represents the number of the basic solutions till the current training epoch. Gradients of Eq.(7) can be calculated by the automatic gradient calculator in Pytorch.

Besides, the values of $R_b$, once being chosen as the source to find the base, no longer participate in approximation error calculation since it can already be expressed by its own basic solution without any error.

## 3. Simulation results and analysis

### 3.1 Simulation Configurations

Task of OOK signal transmission whose bit rates range from 2Gbps to 50Gbps is utilized to demonstrate the performance of the model. Table 1 clearly shows the simulation configurations for the task. All parameters of the transmission link take value from the standard single mode fiber.

**Table 1 PARAMETER CONFIGURATIONS**

| PARAMETER | RANGE | POINTS |
|---|---|---|
| $t$ | [-1, 1] | 312 |
| $\zeta$ | [0, 1] | 11 |
| INITIAL CONDITIONS | OOK with different $R_b$ | 100 |
| $R_b$ | [2Gbps, 50Gbps] | 91 |
| $\alpha$ | $4.605 \times 10^{-5}$ /m | 1 |
| $\beta_2$ | $-2 \times 10^{-26}$ s$^2$/m | 1 |
| $\beta_3$ | $-2 \times 10^{-38}$ s$^3$/m | 1 |
| $n_2$ | $2.6 \times 10^{-20}$ m$^2$/W | 10 |
| $A_{eff}$ | $8 \times 10^{-11}$ m$^2$ | 1 |
| $\lambda$ | 1.55$\mu m$ | 1 |

When it comes to the configuration of the principle driven fiber model as is shown in Table 2, for single pulse predicting, the scale of hidden layers is determined to be 6 layers, each with 100 neurons. Early stopping and maximum training epochs is set to be $10^{-4}$ and 40000 respectively. As for the configurations of the reduced basis expansion method, the maximum number of basis taken for estimating the universal solution is decided to be 12.

**Table 2 MODEL CONFIGURATIONS**

| | TYPE | FULLY CONNECTED |
|---|---|---|
| EACH PRINCIPLE FIBER MODEL | LAYER | [2,100,100,100,100,100,100,2] |
| | NONLINEAR FUNCTION | Tanh($\cdot$) |
| | STOP CRITERION | Loss lower than $10^{-4}$ |
| | | Epochs larger than 40000 |
| REDUCED BASIS | NUMBER OF BASIC SOLUTIONS | 12 |

### 3.2 Training of the Principle Driven Fiber Model

The training algorithm for the previously proposed principle driven fiber model is adaptive gradient descend method (ADAM) [15]. Fig. 2 shows both convergence performance and final prediction of each basic solution. As can be seen from the left column, the loss function descends quickly in the first 1000-2000 epochs. Then it slows down to better search for the local minima. The whole trend of the convergence curve is in consistent with the characteristics of ADAM. Apart from the convergence rate, there still exists several fluctuations which indicates that it can be relatively challenging for the model to map both time and distance coordinate into waveform transmitted.

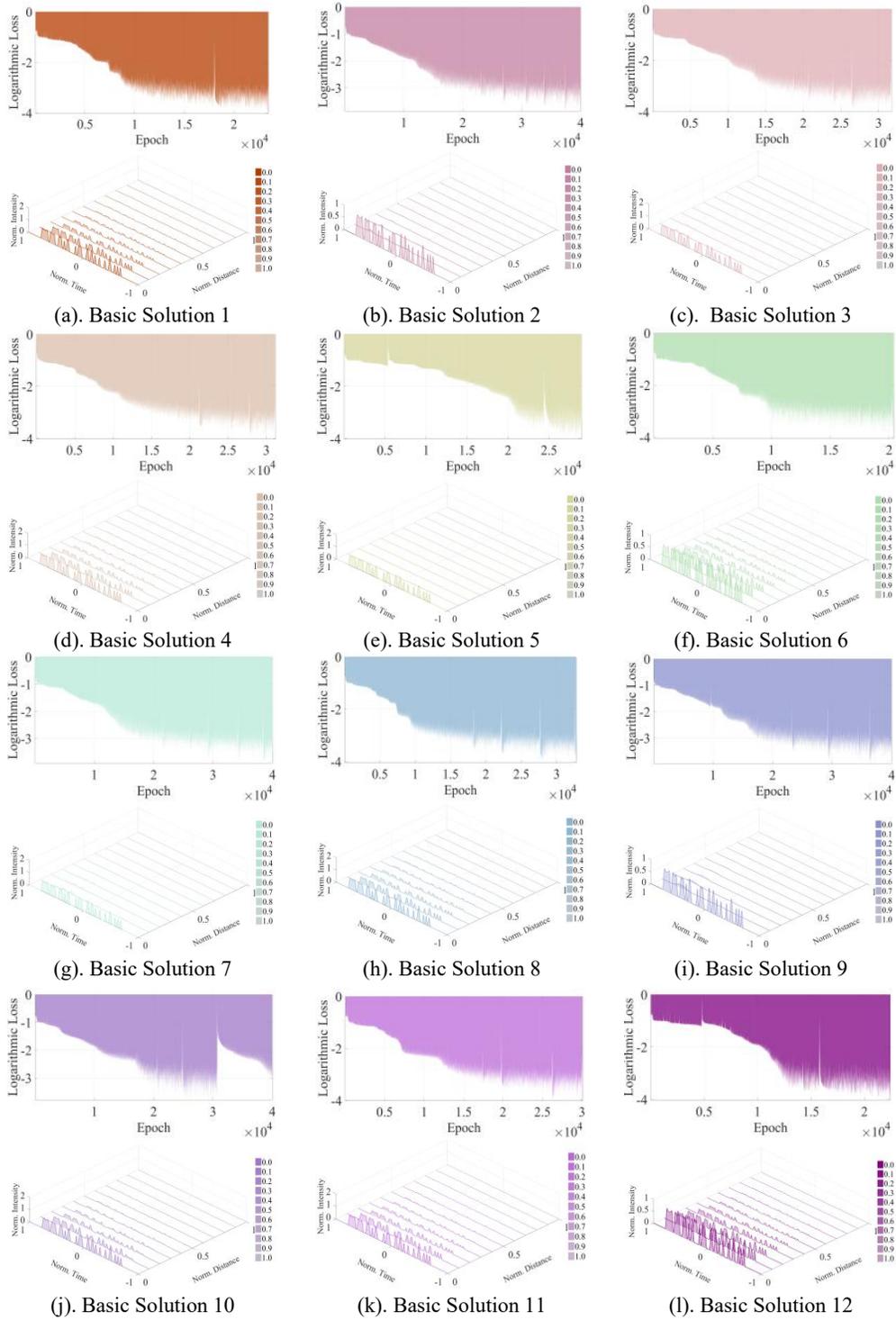

(a). Basic Solution 1

(b). Basic Solution 2

(c). Basic Solution 3

(d). Basic Solution 4

(e). Basic Solution 5

(f). Basic Solution 6

(g). Basic Solution 7

(h). Basic Solution 8

(i). Basic Solution 9

(j). Basic Solution 10

(k). Basic Solution 11

(l). Basic Solution 12

Fig. 2  Principle of parameterized fiber communication model

*3.3 Training of the Reduced Basis Expansion*

After each basic solution is found by the previously proposed fiber transmission model, approximation errors should be minimized by finding of the most appropriate linear combination coefficients with all basic solutions till now and checked for the prediction precision of each different value of $R_b$. Fig.3 clearly shows the process of finding the best linear combination coefficients and the error for each $R_b$.

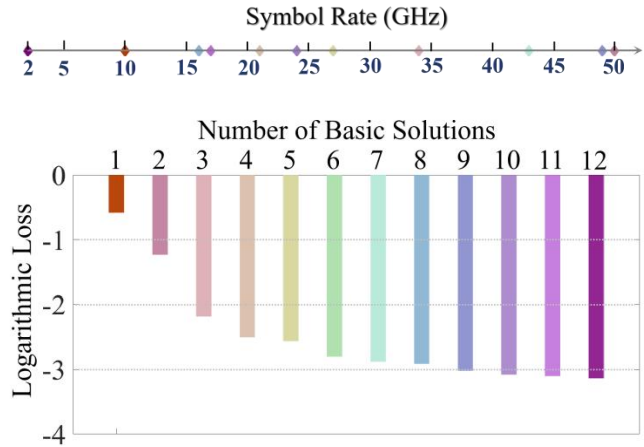

Fig. 3 Basic solution's parameters and the convergence of the reduced basis expansion method

As can be observed from Fig.3, in total 12 basic solutions form as the basis. As the number of basis increases, the total loss function value decreases. Compared with Table I, interesting conclusions can be drawn. New basis surged from the boundary most of the time. It possibly because boundary bit rate like 2GHz and 50GHz differ the most among other values. It can also be indicated from Fig.2 that the value of loss function does not decrease much when adding more basis when the number of basis is larger than 10.

*3.4 Performances of the Parameterized Fiber Model*

The performances of the model is to utilized all basic solutions obtained to better approximate generalized solutions with respect to different $R_b$. Some results for typical values of $R_b$ is shown in Fig.4. What is interesting is that, the basic solutions found for approximation gather around the larger part of bit rates. This phenomenon indicates that the model tends to predict more accurately for the signals with lower bit rates.

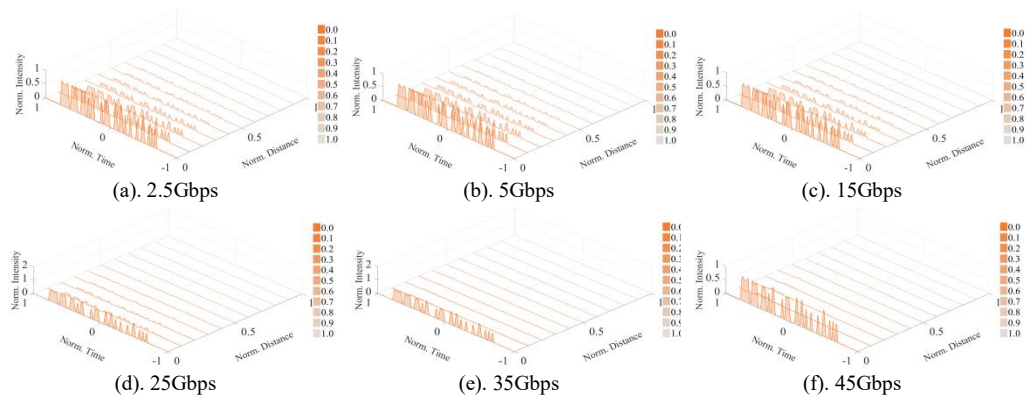

(a). 2.5Gbps  (b). 5Gbps  (c). 15Gbps
(d). 25Gbps  (e). 35Gbps  (f). 45Gbps
Fig. 4. Final performances of the model

The improvement of prediction accuracy can lay on the increase in the scale of the model, i.e, the number of layers, the number of neurons in the previously proposed fiber model, maximum number of basis in the reduced basis expansion method and so on. This is because larger scale of model tend to have better ability of capturing more detailed and rapidly changed characteristics which will results in the increase in prediction accuracy.

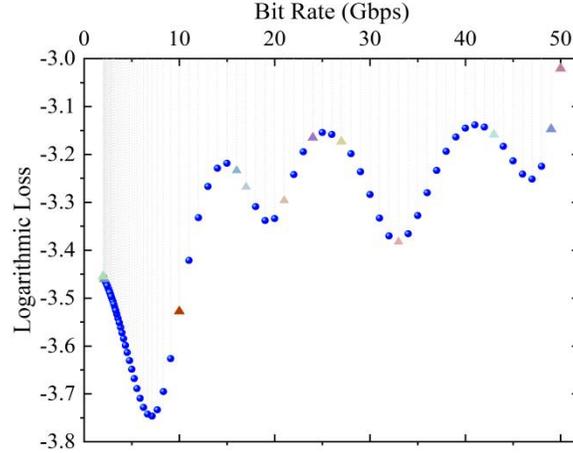

Fig. 5. Final prediction error of the model

Logarithmic loss of the model's prediction with respect to the whole bit rate can be directly seen in Fig. 5. Those bit rates which are chosen to be basic solutions are marked with colored triangles. The color is in consist with that of Fig. 2. Other parameters are marked in blue. As can be seen from the figure, all loss are lower than $10^{-3}$ which shows relatively great prediction accuracy of the model. Those errors of the 'specially selected' parameter to form basic solutions are no longer to be the largest among all except 50Gbps. This is because waveforms with the largest bit rate contains more severe changes during the propagation which becomes the most challenging prediction task for the model.

*3.5 Computational Complexity Comparison*

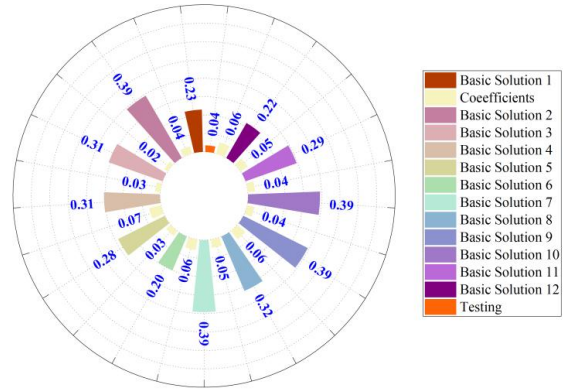

Fig. 6 Computing time of parameterized fiber communication model

Time consumption distribution can be obviously seen in Fig.6. Here, training of basic solutions takes up most of the time. Different basic solution training consumes different time due to the different difficulty of letting the solution converge under different bit rates.

Comparison of computational complexity between SSFM, previously proposed fiber model and the fiber model proposed in this manuscript is also necessary. Few assumptions ought to be put forward before comparison. In order to erase out the differences between each operating software, the scale of multiplications and additions (MAC) is utilized as the comparison standard. Here, there are in total $T$ different values of $R_b$. Both coordinates of time and distance are discretized into $M_t$ and $M_\zeta$ respectively. The longest transmission distance is $L_{max}$. For comparison fairness, both principle driven fiber model and the inserted fiber model inside the parameterized model posses $K$ hidden layers, each layer contains $P$ neurons. For the parameterized fiber model, the maximum basis is determined to be $N_b$. The length of computational unit equal $L_u$. Under these assumptions, the computational complexity of SSFM can be calculated as

$$C_{SSFM} = O\left( \frac{TL_{max}}{L_u} \left[ 4M_t \log_2 M_t + N_{dispersion} + N_{nonlinear} \right] \right) \quad (7\text{-}A)$$

The complexity of the previously proposed fiber model should be

$$C_F = O\left( 2TP + T(K-1)P^2 \right) \quad (7\text{-}B)$$

And the complexity of the parameterized fiber model equals

$$C_{PF} = O\left( 2N_b P + N_b (K-1)P^2 + 2N_b \right) \quad (7\text{-}C)$$

Since $N$ is greatly larger than $T$, it can be easily concluded that the computational complexity of the parameterized fiber model is only the $T/N$ of that of the previously proposed fiber model.

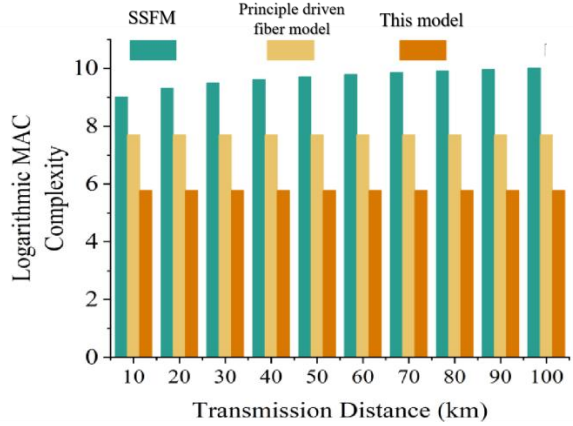

Fig. 7 MAC complexity comparisons

Fig. 7 shows the complexity of the above three different computation complexity. As can be vividly seen, different from SSFM, the complexity of both principle transmission fiber model and generalized fiber transmission model proposed does not increase with the transmission distances. At 100km transmission distance, the complexity of the model proposed is approximately 1% of the previously proposed principle transmission model and 0.01% of SSFM which effectively demonstrates the model's advantages in computing efficiency.

## 4. Conclusions and discussions

In this manuscript, the principle driven fiber model for parameterized input signals is proposed. By transforming the parameter from signals to NLSE, using both fiber model previously proposed and reduced basis expansion method , any solutions with respect to different bit rates can be expressed in a relatively high accuracy.

Several great advantages can be obtained when taking the model into use. First, this model can keep the better balance between prediction accuracy and physical backgrounds since it takes NLSE to design its loss function. Second, compared with other data driven models, this model can still be effectively trained without the needs to collect large scale of transmitted signals before hand. Most importantly, this model can solve signals transmitted at different bit rates all without the need to retrain the whole model which can greatly save large amounts of model's complexity.

**Funding.** the National Key R&D Program of China (2022YFB2903600); the Youth Fund of the National Natural Science Foundation (NSFC) of China under Grant 62301275; Key Laboratory of Radar Imaging and Mircowave Photonics (Nanjing University of Aeronautics and Astronautics), Ministry of Education under Grant NJ20230003; the Research Start-up Fund of Nanjing University of Posts and Telecommunications under Grant NY223032.

**Disclosures.** The authors declare no conflicts of interests.

**Data availability.** Data can be available under reasonable requests.